\documentclass[11pt,a4paper]{article}
\usepackage[hyperref]{acl2018}
\usepackage{times}
\usepackage{latexsym}
\usepackage{graphicx}

\usepackage{url}

\aclfinalcopy % Uncomment this line for the final submission
\title{Financial Aspect-Based Sentiment Analysis using Deep Representations}

\author{Steve Yang \\
  {\tt steve.yang@berkeley.edu} \\\AND
  Jason Rosenfeld \\
  {\tt jrosenfeld@berkeley.edu} \\\AND
  Jacques Makutonin \\
  {\tt jacquesmakutonin@berkeley.edu} \\}

\date{}
\begin{document}
\maketitle
\begin{abstract}
The topic of aspect-based sentiment analysis (ABSA) has been explored for a variety of industries, but it still remains much unexplored in finance. The recent release of data for an open challenge (FiQA) from the companion proceedings of WWW '18 has provided valuable finance-specific annotations. FiQA contains high quality labels, but it still lacks data quantity to apply traditional ABSA deep learning architecture. In this paper, we employ high-level semantic representations and methods of inductive transfer learning for NLP. We experiment with extensions of recently developed domain adaptation methods and target task fine-tuning that significantly improve performance on a small dataset. Our results show an 8.7\% improvement in the F1 score for classification and an 11\% improvement over the MSE for regression on current state-of-the-art results.
\end{abstract}

\section{Introduction}

Aspect-based sentiment analysis (ABSA) is a way to systematically mine opinions given a body of text. Unlike regular sentiment analysis, ABSA allows for far more granular levels of opinion mining. For example, one common application of ABSA is to dissect product or service reviews and determine sentiment on sometimes unrelated aspects, such as a quality or price. Rarely are product reviews as simple as good or bad. They are nuanced with conflicting positive and negative opinions based on what aspect of the product is being reviewed.

We find the field of finance to be a significantly under-explored domain for ABSA. Similar to product reviews, financial investment opportunities are commonly written as free-form essays; these write-ups are generally nuanced with positive and negative opinions on specific aspects of a certain investment opportunity. Being able to identify these topics and to subsequently determine the associated sentiment could be beneficial in downstream models to auto-summarize predefined aspects, allowing users to obtain structured information from an unstructured set of write-ups. Another use case could be to employ the aspect-based sentiments as features to classify future performance or volatility of investment ideas.

The under-explored nature of financial related ABSA also manifests itself in a lack of large, high-quality datasets on which to train. Current ABSA techniques and model architectures do not accurately scale down to small data sizes, presenting an opportunity for transfer learning that can leverage larger, domain-related datasets. Successful inductive transfer learning allows these larger datasets that have no sentiment or aspect annotations to be used to improve results on the main ABSA task.

\section{Related Work}

ABSA is not a new idea. There have been extensive bodies of work, starting from the original rule-based methods \cite{thet2010aspect} to more recent deep learning methods \cite{wang2015deep}. In general, the state-of-the-art consists of a few subtasks. First, a model identifies the entity and its aspects. Then, sentiment analysis is performed on the body of text before combining the two tasks. 

Sentiment analysis for finance -- without considering aspects -- has also been explored \cite{cortis-EtAl:2017:SemEval}. Unlike ABSA, more general sentiment analysis cannot determine if a statement is positive or negative by each aspect -- it must choose an overall sentiment.

Until recently, it has been difficult to perform the kind of analysis in \newcite{wang2015deep} in the financial domain due to both a lack of well-annotated datasets and the necessary quantity of data. Unlike the larger product review dataset \cite{pontiki2016semeval} used in Wang's study, we speculate it would be expensive to annotate as many examples for the financial domain, as it would require extensive domain expertise. However, as part of the companion proceedings for WWW '18 conference, \newcite{maia201818} released a very small dataset (FiQA) with a call for papers. FiQA contains the particular labels for which we are interested, but it lacks data quantity. Still, we reference the submissions to this open challenge as a response to FiQA Task 1. Many of the submissions \cite{jangid2018aspect, Chen:2018, deFrancaCosta:2018} use a neural architecture similar to \newcite{wang2015deep}, despite very few training examples. We show these results along with our own in Section~\ref{ssec:results}.

The problem of few training examples brings us to the topic of inductive transfer learning \cite{pan2010survey}. In general, transfer learning allows us to perform training on some source task with the ultimate goal of optimizing the loss of some target task. Word embeddings such as word2vec \cite{mikolov2013distributed} or GloVe \cite{pennington2014glove} are an early form of transfer learning in NLP. Just beyond that are more sophisticated vector representations of words, such as ELMo \cite{peters2018deep}. ELMo embeddings offer a solution to the challenges of complex word use across linguistic context (i.e. polysemy, syntax, etc.) and limited training data. In ELMo, each word is assigned a representation which is a function of the entire corpus sentences to which they belong. At a high-level, the embeddings are computed from internal states of a two-layer biLSTM, bidirectional Language Model (biLM). Despite ELMo's important improvements on ``traditional" word embeddings, any approach utilizing ELMo embeddings still requires a custom architecture further downstream. 

In this paper, we explore a recent method called ULMFiT \cite{howard2018universal}, which allows us to not only transfer word representations with a vector, but use a single, pre-trained model architecture (AWD-LSTM \citealp{DBLP:journals/corr/abs-1708-02182}) for all intended tasks. Similar to ELMo, the key benefit of a higher level representation is that it allows semantically meaningful starting points of the input words for training. Using ELMo, one ultimately concatenates the output of each trained layer and uses it as a fixed embedding for some downstream task, whereas ULMFiT fine-tunes an entire language model to some target domain and then directly connects a downstream target task. This concept itself is not entirely new \cite{DBLP:journals/corr/DaiL15a}, but \newcite{howard2018universal} contribute novel techniques (Gradual Unfreezing, Discriminative Learning Rates and Slanted Triangular Learning Rates) to make this possible on small datasets without all prior learning being forgotten. Similar to chain-thaw \cite{felbo2017using}, gradual unfreezing offers another approach to the transfer learning training and fine-tuning process. Chain-thaw  first tunes any new layers in a model until convergence; this is followed by a sequential tuning of each layer individually. Finally, chain-thaw fine-tunes all layers together. In contrast, gradual unfreezing fine-tunes all layers in reverse, adding a 'thawed' layer instead of fine-tuning single layers individually.

\section{Contributions}

The following are the primary contributions of this paper:

\begin{itemize}
  \item We assess the performance of recent NLP inductive transfer learning techniques such as ULMFiT on a new dataset (FiQA) which has the problem of limited training examples.
  \item We use FiQA as our target task to conduct experiments with varying intermediary tasks, at least one of which appears to be novel, to our knowledge.
  \item We extend state-of-the-art on FiQA task 1 (aspect 2 classification and sentiment scoring) by using our novel intermediary task.
\end{itemize}

The rest of this paper is structured as follows. In Section~\ref{sec:data} we introduce the critical datasets used for both the primary and intermediary tasks (outlined in Section~\ref{sec:tasks}). We propose our novel variation to model architecture and hyper-parameter tuning in Section~\ref{sec:methods}. In Section~\ref{sec:results} we discuss our results and analysis of our experiments. Finally, we conclude with some future direction of research related to this paper in Section~\ref{sec:conclusion}.

\section{Data}
\label{sec:data}

\subsection{FiQA}
The provided training dataset for WWW '18 \cite{maia201818} contains a total of 1,174 examples from news headlines and tweets. Each example contains the sentence and the sentence snippet associated with the target entity, aspect, and sentiment score. A sample FiQA entry is shown in Table~\ref{fiqa-ex}.

\begin{table}
\begin{center}
\begin{tabular}{|p{3cm}|p{3cm}|}
\hline
\bf Sentence & easyJet expects resilient demand to withstand security fears. \\
\hline
\bf Aspect Level 1 & Corporate \\
\hline
\bf Aspect Level 2 & Risks \\
\hline
\bf Sentiment Score & 0.165 \\
\hline
\bf Target & easyJet \\
\hline
\end{tabular}
\end{center}
\label{fiqa-ex}
\caption{ An example entry from FiQA }
\end{table}

A Level 1 Aspect label takes on one of four possible labels (Corporate, Economy, Market or Stock), and our Level 2 Aspect label takes on one of  twenty-seven possible labels (Appointment, Risks, Dividend Policy, Financial, Legal, Volatility, Coverage, Price Action, etc.). The original dataset contained a small number of multilabel examples, however, we considered this number too few to train a meaningful multilabel classifier. Thus, we slightly stray from the original WWW '18 task for the purpose of this research. Finally, sentiment score takes on a continuous value between -1 and 1 -- most negative to most positive.

\subsection{Value Investors Club (VIC)}
VIC is an online investment forum where fund managers and skilled investors submit in-depth investment recommendations on a daily basis.  We scraped roughly 6,800 documents with investment theses along with attributes of a particular thesis, such as long or short stock position. Each document has an average of approximately 1,950 words which gives us a corpus size of about 13M. This is a crucial dataset that we use to adapt the domain of our general pre-trained models to our tasks in FiQA subsection~\ref{ssec:ulmfit}. We find the high quality of VIC to be of particular interest. Although most recommendations and investment ideas shared online lack credibility or accuracy, VIC investment ideas empirically outperform the market on average and over time \cite{gray2012fund}.

\subsection{Pre-Trained Models}
Although we do not directly train on ``1B Word Benchmark" corpus \cite{chelba2013one} or wikitext-103 \cite{merity2016pointer}, we thought it was important to note the underlying source for the pre-trained models released by \newcite{howard2018universal} and \newcite{peters2018deep}.

\section{Tasks}
\label{sec:tasks}
\begin{figure*}
\includegraphics[width=\textwidth,height=5.5cm]{{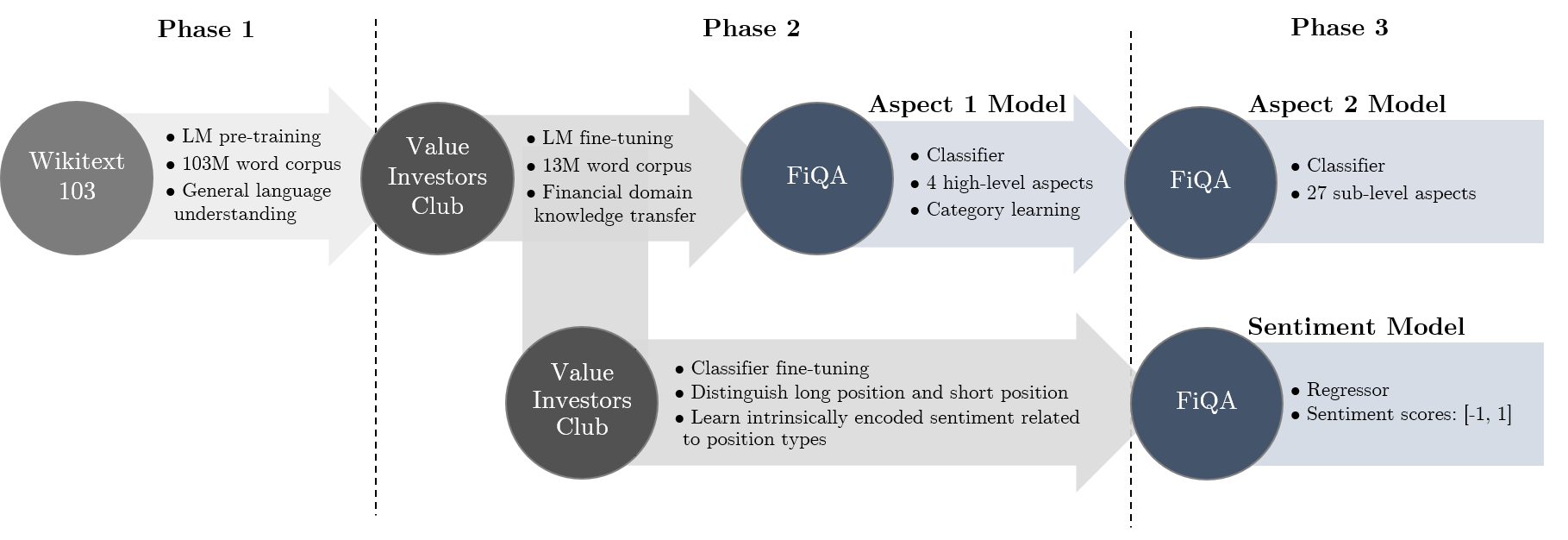}}
\caption{Our custom ULMFiT training process consists of several stages: for all final tasks, our base models begin with a language model pretrained on Wikitext-103 followed by a fine-tuning step trained on the VIC corpus. For further steps, the language model's decoder is removed and replaced with a linear head. FiQA Aspect Level 1 classification is first trained before then passing that same model onto training for FiQA Aspect Level 2 classification. For our regression task, the fine-tuned language model performs an intermediary classification task on known attributes of the VIC write-ups (namely, whether a write-up advocates for a long or short position). Lastly, the model is trained as a regressor on FiQA sentiment scores.}
\label{fig:methods}
\end{figure*}

Using this outlined data, we have two primary subtasks: classification for determining sentence aspects and regression for continuous sentiment. Namely, FiQA Level 2 Aspect and FiQA sentiment score are our ultimate target values of interest. We measure our models by using error-rate and F1-score for the classification task, and with MSE and R-squared for the regression task. These tasks can be seen in Figure~\ref{fig:methods}

We also have a number of intermediary tasks, the purpose of which is to better adapt the pre-trained models to our primary tasks. First, we use VIC to train a language model on the 13M word corpus. We also train VIC on a binary classifier which learns if the thesis of a given body of text is that of a long position or short position. We also use FiQA Level 1 Aspect as an intermediary task.

\section{Methods}
\label{sec:methods}

\subsection{Baseline}
Our most naive baseline on the performance of the primary tasks are logistic regression for aspect classification and linear regression for sentiment. We use a simple sparse representation of words, rather than introduce any type of continuous embedding. 

We also create a simple ELMo embedding baseline from the \newcite{peters2018deep} pre-trained biLM. Our model computes a fixed mean-pooling of all contextualized word representations for each input sentence which is then passed through a single dense layer. Note that we do not claim this methodology to be comparable to our more rigorous implementation of ULMFiT. We use this as a neural baseline, but reserve any in-depth analysis for future work.

\subsection{ULMFiT}
\label{ssec:ulmfit}

We largely use the methodology and architecture used in the ULMFiT \cite{howard2018universal} paper and experiment with different methods of domain adaptation, model fine-tuning, and hyper-parameter tuning.

Typically, ULMFiT is composed of three phases of training. First, a language model is trained using a large general corpus. In particular, \newcite{howard2018universal} released a pre-trained general language model which we use as our starting point. Second, the weights from the first phase are fine-tuned to a general task using a corpus from the target domain. Third, the weights from the second phase are fine-tuned to the primary task using the same upstream architecture as the first two phases, but by passing the output of the tuned LSTM model into a new fully-connected layer.

We propose a few different variations of the typical framework and apply it to FiQA:

\begin{enumerate}
\item FiQA does not contain enough volume of data to meaningfully apply LM fine-tuning (Phase 2); Thus, we use VIC as our target corpus for the language model fine-tuning. In this phase, we apply no change in model architecture.
\item We further exploit VIC labels of long/short positions and perform an additional round of fine-tuning in Phase 2. The decoder of the Language Model is no longer used, and, in its place, we add a fully-connected layer with a binary output.
\item We train the FiQA primary classification and regression tasks using both variations of phase two, attaching classification and regression fully-connected layers, respectively.
\item We train FiQA Aspect Level 1 using both variations of phase two, then transfer the weights further downstream for FiQA Aspect Level 2.
\end{enumerate}

\subsection{Hyper-parameters}
As suggested in \newcite{howard2018universal}, there are a number of techniques used to ensure that the training process during fine-tuning does not cause the model to ``forget`` what was previously learned. We experiment with both gradual unfreezing \shortcite{howard2018universal} and chain-thaw \shortcite{felbo2017using} and compare results on varying sub-sample sizes. We also experiment with some early-stoppage of chain-thaw, which we think may be beneficial to avoid catastrophic forgetting and to not overly fine-tune. In addition, we utilize slanted triangular learning rates, concat pooling, and bptt for text classification as described in \newcite{howard2018universal}. While we did initially perform some exploratory work on different hyper-parameter values of these latter techniques, we keep the values constant for our experiments and results.

\begin{table*}
\includegraphics[width=\textwidth,height=5.7cm]{{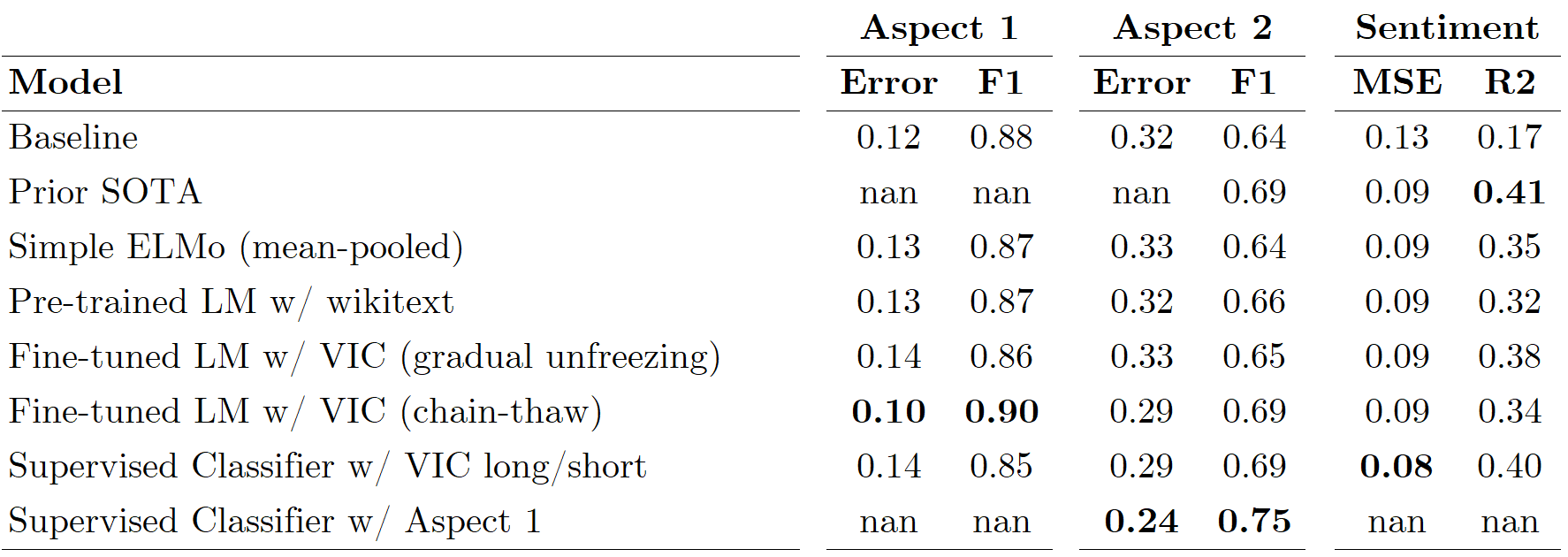}}
\caption{Test error rates and F1 scores for classification, test MSE and R2 for regression.}
\label{tab:results}
\end{table*}

\section{Results and Discussions}
\label{sec:results}

\subsection{Model Results}
\label{ssec:results}
We were able to ultimately outperform the current FiQA Task 1 state-of-the-art by using the ULMFiT framework in conjunction with our modified intermediary tasks (see Table~\ref{tab:results}). In some cases we achieve close to state-of-the-art scores, even without much fine-tuning.

In Aspect Level 2 classification, although we can match state-of-the-art results with a typical ULMFiT process, we are unable to exceed state-of-the-art until we use Aspect Level 1 as our intermediary task. We conclude that since Aspect Level 1 and Level 2 are hierarchical, the internal states from fine-tuning on Aspect Level 1 lead to a much better starting point for training on Aspect Level 2. We want to be clear that this method does not train with Aspect Level 1 as a feature, we simply transfer the internal state. Thus, it will work at inference time without Aspect Level 1 labels. We also believe that this is not simply a result of additional training as the prior metrics for Aspect Level 2 had already converged. 

While Aspect Level 1 was not our primary task, we note some interesting results here as well. We saw the best performance from using chain-thaw. For this particular task, we see that the other models actually perform worse than using the Phase 1 language model without any domain adaptation. It is possible that these methods of training here are leading to catastrophic forgetting \cite{howard2018universal}, in which we lose the utility of the pre-trained language model.

For the regression task, we outperform the current FiQA state-of-the-art in terms of MSE, but not in terms of R-squared. It is interesting to note that the top-performer was from Phase 2 training on whether or not VIC advocates for a long or short position. It seems rational that transfer learning from a more closely related task leads to better starting internal states for our primary sentiment task.

We also note that in all cases, even our Phase 1 language model is comparable, if not superior, to our naive baseline models. In this case, no additional training time is required to produce these predictions. This framework is also relatively universal in the sense that not much pre-preprocessing is required other than standard tokenization. Out-of-vocabulary words are also easily handled by using the mean representation of all other vocabulary.

\subsection{Hyper-parameter results}

\begin{figure*}
\includegraphics[width=\textwidth,height=5.7cm]{{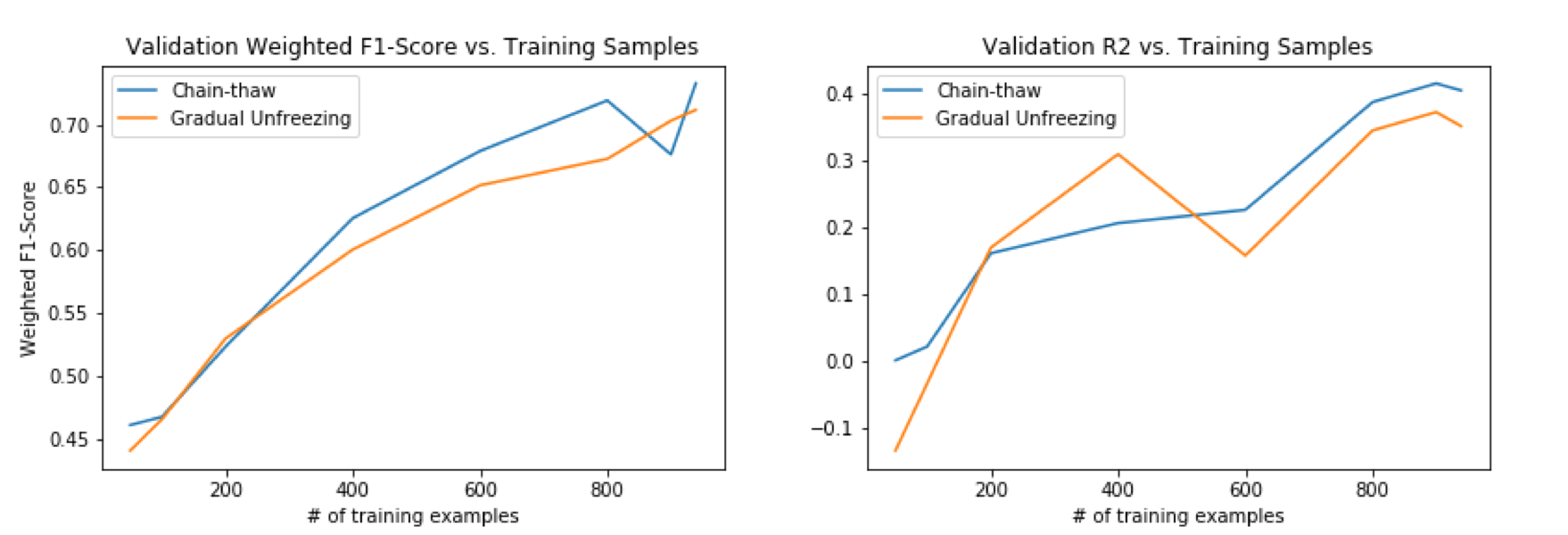}}
\caption{Validation F1-scores on Aspect 2 classification and R2 on sentiment regression across a range of training examples. We find chain thaw to be a more stable and predictable learning process at smaller training sample sizes.
}
\label{fig:subsample}
\end{figure*}

The only hyper-parameter we vary for each of the models above is the method in which layers are unfrozen. We report gradual unfreezing for all models, but only report the best version of chain-thaw for each model for brevity. Roughly speaking, our primary tasks perform the best with full chain-thaw, but our Aspect Level 1 classification performs best with partial chain-thaw. More research is needed on this topic, but we speculate that this may be due to over-learning on the target task.

Also, we compare model performance for both primary tasks with random subsamples of the training data (see Figure~\ref{fig:subsample}). We see that chain-thaw tends to have a smoother decrease in error. It appears the steepest portion of the curve occurs when there are even fewer training examples than our case. While it's difficult to further forecast the error rates, this shows some evidence our current methodology does help with the few training examples issue. However, the error rates do not yet appear to have hit a point of saturation.

\section{Conclusion}
\label{sec:conclusion}
Aspect-based sentiment analysis is not broadly used in the finance industry. Due to a lack of well annotated data, it is necessary to apply transfer learning techniques to leverage data from a large general corpus for the training of tasks on specific domain data.

Because the dataset for ABSA training (FiQA) is so small, we modify the ULMFiT steps. We fine tuned the language model with another dataset containing target domain specific context (VIC) and perform additional fine-tuning on a classification task (long/short position of VIC documents). Another contribution is the improvement of the FiQA Aspect 2 classification task by leveraging a pre-trained model on Aspect 1 classification.

Despite improved performance using a modified ULMFiT process for ABSA training on a very small sample, more research is needed to generalize this methodology to multi-task and multi-label learning. Moreover, in order to improve the methodology, experiments using other pre-trained models of large corpora as well as different setups of hyper-parameters should be evaluated.

The adapted ULMFiT methodology to ABSA on a very small, domain-specific dataset such as FiQA is an important cornerstone for learning subsequent tasks related to predicting financial performance. The methodology can also be expanded to other domain specific tasks when the size of data available represents a challenge.

%\bibliography{acl2018}
\nocite{*}
\bibliographystyle{acl_natbib}
\bibliography{report}

\begin{thebibliography}{}
\expandafter\ifx\csname natexlab\endcsname\relax\def\natexlab#1{#1}\fi

\bibitem[{Blitzer et~al.(2007)Blitzer, Dredze, and
  Pereira}]{blitzer2007biographies}
John Blitzer, Mark Dredze, and Fernando Pereira. 2007.
\newblock Biographies, bollywood, boom-boxes and blenders: Domain adaptation
  for sentiment classification.
\newblock In {\em Proceedings of the 45th annual meeting of the association of
  computational linguistics\/}. pages 440--447.

\bibitem[{Blitzer et~al.(2006)Blitzer, McDonald, and
  Pereira}]{blitzer2006domain}
John Blitzer, Ryan McDonald, and Fernando Pereira. 2006.
\newblock Domain adaptation with structural correspondence learning.
\newblock In {\em Proceedings of the 2006 conference on empirical methods in
  natural language processing\/}. Association for Computational Linguistics,
  pages 120--128.

\bibitem[{Chelba et~al.(2013)Chelba, Mikolov, Schuster, Ge, Brants, Koehn, and
  Robinson}]{chelba2013one}
Ciprian Chelba, Tomas Mikolov, Mike Schuster, Qi~Ge, Thorsten Brants, Phillipp
  Koehn, and Tony Robinson. 2013.
\newblock One billion word benchmark for measuring progress in statistical
  language modeling.
\newblock {\em arXiv preprint arXiv:1312.3005\/} .

\bibitem[{Chen et~al.(2018)Chen, Huang, and Chen}]{Chen:2018}
Chung-Chi Chen, Hen-Hsen Huang, and Hsin-Hsi Chen. 2018.
\newblock \href{https://doi.org/10.1145/3184558.3191824}{Fine-grained analysis
  of financial tweets}.
\newblock In {\em Companion Proceedings of the The Web Conference 2018\/}.
  International World Wide Web Conferences Steering Committee, Republic and
  Canton of Geneva, Switzerland, WWW '18, pages 1943--1949.
\newblock
  \href{https://doi.org/10.1145/3184558.3191824}{https://doi.org/10.1145/3184558.3191824}.

\bibitem[{Cortis et~al.(2017)Cortis, Freitas, Daudert, Huerlimann, Zarrouk,
  Handschuh, and Davis}]{cortis-EtAl:2017:SemEval}
Keith Cortis, Andr\'{e} Freitas, Tobias Daudert, Manuela Huerlimann, Manel
  Zarrouk, Siegfried Handschuh, and Brian Davis. 2017.
\newblock \href{http://www.aclweb.org/anthology/S17-2089}{Semeval-2017 task 5:
  Fine-grained sentiment analysis on financial microblogs and news}.
\newblock In {\em Proceedings of the 11th International Workshop on Semantic
  Evaluation (SemEval-2017)\/}. Association for Computational Linguistics,
  Vancouver, Canada, pages 519--535.
\newblock
  \href{http://www.aclweb.org/anthology/S17-2089}{http://www.aclweb.org/anthology/S17-2089}.

\bibitem[{Dai and Le(2015)}]{DBLP:journals/corr/DaiL15a}
Andrew~M. Dai and Quoc~V. Le. 2015.
\newblock \href{http://arxiv.org/abs/1511.01432}{Semi-supervised sequence
  learning}.
\newblock {\em CoRR\/} abs/1511.01432.
\newblock
  \href{http://arxiv.org/abs/1511.01432}{http://arxiv.org/abs/1511.01432}.

\bibitem[{de~Fran\c{c}a~Costa and da~Silva(2018)}]{deFrancaCosta:2018}
Dayan de~Fran\c{c}a~Costa and Nadia Felix~Felipe da~Silva. 2018.
\newblock \href{https://doi.org/10.1145/3184558.3191828}{Inf-ufg at fiqa 2018
  task 1: Predicting sentiments and aspects on financial tweets and news
  headlines}.
\newblock In {\em Companion Proceedings of the The Web Conference 2018\/}.
  International World Wide Web Conferences Steering Committee, Republic and
  Canton of Geneva, Switzerland, WWW '18, pages 1967--1971.
\newblock
  \href{https://doi.org/10.1145/3184558.3191828}{https://doi.org/10.1145/3184558.3191828}.

\bibitem[{Felbo et~al.(2017)Felbo, Mislove, S{\o}gaard, Rahwan, and
  Lehmann}]{felbo2017using}
Bjarke Felbo, Alan Mislove, Anders S{\o}gaard, Iyad Rahwan, and Sune Lehmann.
  2017.
\newblock Using millions of emoji occurrences to learn any-domain
  representations for detecting sentiment, emotion and sarcasm.
\newblock {\em arXiv preprint arXiv:1708.00524\/} .

\bibitem[{Gray et~al.(2012)Gray, Crawford, and Kern}]{gray2012fund}
Wesley~R Gray, Steve Crawford, and Andrew~E Kern. 2012.
\newblock Do fund managers identify and share profitable ideas? .

\bibitem[{Howard and Ruder(2018)}]{howard2018universal}
Jeremy Howard and Sebastian Ruder. 2018.
\newblock Universal language model fine-tuning for text classification.
\newblock In {\em Proceedings of the 56th Annual Meeting of the Association for
  Computational Linguistics (Volume 1: Long Papers)\/}. volume~1, pages
  328--339.

\bibitem[{Jangid et~al.(2018)Jangid, Singhal, Shah, and
  Zimmermann}]{jangid2018aspect}
Hitkul Jangid, Shivangi Singhal, Rajiv~Ratn Shah, and Roger Zimmermann. 2018.
\newblock Aspect-based financial sentiment analysis using deep learning.
\newblock In {\em Companion of the The Web Conference 2018 on The Web
  Conference 2018\/}. International World Wide Web Conferences Steering
  Committee, pages 1961--1966.

\bibitem[{Maia et~al.(2018)Maia, Handschuh, Freitas, Davis, McDermott, Zarrouk,
  and Balahur}]{maia201818}
Macedo Maia, Siegfried Handschuh, Andr{\'e} Freitas, Brian Davis, Ross
  McDermott, Manel Zarrouk, and Alexandra Balahur. 2018.
\newblock Www'18 open challenge: Financial opinion mining and question
  answering.
\newblock In {\em Companion of the The Web Conference 2018 on The Web
  Conference 2018\/}. International World Wide Web Conferences Steering
  Committee, pages 1941--1942.

\bibitem[{Merity et~al.(2017)Merity, Keskar, and
  Socher}]{DBLP:journals/corr/abs-1708-02182}
Stephen Merity, Nitish~Shirish Keskar, and Richard Socher. 2017.
\newblock \href{http://arxiv.org/abs/1708.02182}{Regularizing and optimizing
  {LSTM} language models}.
\newblock {\em CoRR\/} abs/1708.02182.
\newblock
  \href{http://arxiv.org/abs/1708.02182}{http://arxiv.org/abs/1708.02182}.

\bibitem[{Merity et~al.(2016)Merity, Xiong, Bradbury, and
  Socher}]{merity2016pointer}
Stephen Merity, Caiming Xiong, James Bradbury, and Richard Socher. 2016.
\newblock Pointer sentinel mixture models.
\newblock {\em arXiv preprint arXiv:1609.07843\/} .

\bibitem[{Mikolov et~al.(2013)Mikolov, Sutskever, Chen, Corrado, and
  Dean}]{mikolov2013distributed}
Tomas Mikolov, Ilya Sutskever, Kai Chen, Greg~S Corrado, and Jeff Dean. 2013.
\newblock Distributed representations of words and phrases and their
  compositionality.
\newblock In {\em Advances in neural information processing systems\/}. pages
  3111--3119.

\bibitem[{Pan et~al.(2010)Pan, Yang et~al.}]{pan2010survey}
Sinno~Jialin Pan, Qiang Yang, et~al. 2010.
\newblock A survey on transfer learning.
\newblock {\em IEEE Transactions on knowledge and data engineering\/}
  22(10):1345--1359.

\bibitem[{Pennington et~al.(2014)Pennington, Socher, and
  Manning}]{pennington2014glove}
Jeffrey Pennington, Richard Socher, and Christopher Manning. 2014.
\newblock Glove: Global vectors for word representation.
\newblock In {\em Proceedings of the 2014 conference on empirical methods in
  natural language processing (EMNLP)\/}. pages 1532--1543.

\bibitem[{Peters et~al.(2018)Peters, Neumann, Iyyer, Gardner, Clark, Lee, and
  Zettlemoyer}]{peters2018deep}
Matthew~E Peters, Mark Neumann, Mohit Iyyer, Matt Gardner, Christopher Clark,
  Kenton Lee, and Luke Zettlemoyer. 2018.
\newblock Deep contextualized word representations.
\newblock {\em arXiv preprint arXiv:1802.05365\/} .

\bibitem[{Pontiki et~al.(2016)Pontiki, Galanis, Papageorgiou, Androutsopoulos,
  Manandhar, Mohammad, Al-Ayyoub, Zhao, Qin, De~Clercq
  et~al.}]{pontiki2016semeval}
Maria Pontiki, Dimitris Galanis, Haris Papageorgiou, Ion Androutsopoulos,
  Suresh Manandhar, AL-Smadi Mohammad, Mahmoud Al-Ayyoub, Yanyan Zhao, Bing
  Qin, Orph{\'e}e De~Clercq, et~al. 2016.
\newblock Semeval-2016 task 5: Aspect based sentiment analysis.
\newblock In {\em Proceedings of the 10th international workshop on semantic
  evaluation (SemEval-2016)\/}. pages 19--30.

\bibitem[{Thet et~al.(2010)Thet, Na, and Khoo}]{thet2010aspect}
Tun~Thura Thet, Jin-Cheon Na, and Christopher~SG Khoo. 2010.
\newblock Aspect-based sentiment analysis of movie reviews on discussion
  boards.
\newblock {\em Journal of information science\/} 36(6):823--848.

\bibitem[{Wang and Liu(2015)}]{wang2015deep}
Bo~Wang and Min Liu. 2015.
\newblock Deep learning for aspect-based sentiment analysis.

\end{thebibliography}

\end{document}